\documentclass{article}

\usepackage{arxiv}

\usepackage[utf8]{inputenc} 
\usepackage[T1]{fontenc}    
\usepackage{hyperref}       
\usepackage{url}            
\usepackage{booktabs}       
\usepackage{amsfonts}       
\usepackage{nicefrac}       
\usepackage{microtype}      
\usepackage{lipsum}
\usepackage{svg}
\usepackage{graphicx}
\graphicspath{ {./images/} }
\usepackage{amsmath} 
\usepackage{amssymb}    
\usepackage{graphicx}    
\usepackage{geometry}    
\usepackage{booktabs}    
\usepackage{pdflscape}
\usepackage{xcolor}
\usepackage[normalem]{ulem}
\usepackage{comment}
 \usepackage{longtable}
\usepackage{tabularx}
\usepackage{caption} 
\usepackage{subcaption}
\usepackage{longtable}
\usepackage{booktabs}
\usepackage{amssymb}
\usepackage[utf8]{inputenc}

\title{YOLOv12: A Breakdown of the Key Architectural Features}

\author{
  \textbf{Mujadded Al Rabbani Alif}\textsuperscript{*} and \textbf{Muhammad Hussain}\\[1ex] 
  \begin{minipage}[t]{0.90\textwidth}
    \scriptsize Department of Computer Science, Huddersfield University, Queensgate, Huddersfield HD1 3DH, UK; \\
    \textsuperscript{*}Correspondence: m.alif@hud.ac.uk;
  \end{minipage}
}
\begin{document}
\maketitle
\begin{abstract}
This paper presents an architectural analysis of YOLOv12, a significant advancement in single-stage, real-time object detection building upon the strengths of its predecessors while introducing key improvements. The model incorporates an optimised backbone (R-ELAN), 7×7 separable convolutions, and FlashAttention-driven area-based attention, improving feature extraction, enhanced efficiency, and robust detections. With multiple model variants, similar to its predecessors, YOLOv12 offers scalable solutions for both latency-sensitive and high-accuracy applications. Experimental results manifest consistent gains in mean average precision (mAP) and inference speed, making YOLOv12 a compelling choice for applications in autonomous systems, security, and real-time analytics. By achieving an optimal balance between computational efficiency and performance, YOLOv12 sets a new benchmark for real-time computer vision, facilitating deployment across diverse hardware platforms, from edge devices to high-performance clusters.
\end{abstract}


\section{Introduction}

Since its inception, the YOLO (You Only Look Once) series has been at the forefront of real-time object detection. It is known for its ability to streamline detection by predicting bounding boxes and class probabilities in a single pass through the network. Introduced by Redmon et al., the YOLO framework has continuously evolved, each iteration building upon its predecessors to enhance speed and accuracy, which are critical for applications ranging from Autonomous Vehicles to Surveillance Systems \cite{redmon2015you}, Agriculture Domain to Vehicle Detection\cite{alif2024yolov1, sundaresan2024comparative, alif2024yolov11} and from Healthcare to Manufacturing \cite{hussain2022gradient}. Significant architectural innovations have marked the progression from YOLOv1 to YOLOv11. YOLOv2 and YOLOv3 expanded the model's capabilities through multi-scale feature extraction layers and more sophisticated training strategies \cite{redmon2016yolo9000betterfasterstronger}. Subsequent versions, such as YOLOv4 through YOLOv6, focused on refining the balance between computational efficiency and detection precision, incorporating techniques like mosaic data augmentation and CSPNet to optimise performance \cite{bochkovskiy2020yolov4optimalspeedaccuracy}.

Later iterations YOLOv7, YOLOv8, and YOLOv9 introduced increased adaptability, enabling robust performance across a spectrum of hardware environments, from constrained edge devices to high-capacity GPUs \cite{wang2021scaledyolov4scalingcrossstage}. YOLOv10 and YOLOv11 further advanced these capabilities by integrating state-of-the-art deep learning methodologies, such as attention mechanisms and transformer-inspired components, improving the model's ability to discern complex visual patterns in diverse scenes \cite{ge2021yoloxexceedingyoloseries}. Despite these advancements, the demand for higher accuracy in detecting small, partially occluded, or overlapping objects, especially under real-time constraints, remained an ongoing challenge \cite{alif2024yolov11vehicledetectionadvancements}.

YOLOv12 represents the latest leap forward, introducing revolutionary architectural enhancements that promise to redefine real-time object detection \cite{tian2025yolov12attentioncentricrealtimeobject}. Building on the strong foundations of its predecessors, YOLOv12 addresses previously unmet needs by leveraging an \emph{attention-centric design} centred on \emph{area attention}, which segments feature maps to focus more effectively on critical regions. This attention module is accelerated by \emph{FlashAttention} to reduce memory overhead, allowing for near real-time processing at high resolutions. In pair, a \emph{Residual Efficient Layer Aggregation Network} (R-ELAN) alleviates gradient bottlenecks and improves feature fusion, while \emph{7$\times$7 separable convolutions} replace traditional positional encodings, preserving spatial context with fewer parameters. These innovations collectively boost detection accuracy, especially for smaller or heavily occluded objects, without compromising the hallmark real-time performance of the YOLO series.

The development of YOLOv12 is motivated by the growing complexity of real-world vision tasks and the push toward deploying advanced deep learning models in resource-constrained environments, including mobile and edge devices. Although YOLOv11 made significant strides in accuracy and adaptability, maintaining high throughput under stringent hardware limitations still proved challenging \cite{alif2024yolov11vehicledetectionadvancements}. YOLOv12 tackles these hurdles by improving computational efficiency through targeted optimisations like FlashAttention, adaptive MLP ratios, and refined convolutional strategies \cite{tian2025yolov12attentioncentricrealtimeobject}. Such refinements reduce the memory footprint and inference latency, making YOLOv12 an attractive option for applications ranging from autonomous navigation, where rapid, accurate object detection is paramount, to embedded vision systems like robots or drones, which operate under tight power and compute constraints.

Beyond its technical innovations, YOLOv12 continues the YOLO tradition of broad applicability. Its enhanced feature extraction capabilities enable more reliable detection in dense environments such as urban traffic and crowded public spaces. In the automotive sector, it can bolster the reliability of advanced driver-assistance systems (ADAS) and autonomous vehicles through more precise detection and tracking of road users. YOLOv12's improved accuracy in healthcare may facilitate detailed analysis of medical images, detecting anomalies in radiological scans or segmenting anatomical structures. Meanwhile, agriculture can benefit from robust small-object detection to monitor crop health and identify pests or diseases early \cite{redmon2016yolo9000betterfasterstronger, bochkovskiy2020yolov4}.

In sum, YOLOv12 is poised to contribute significantly to computer vision by delivering notable improvements in speed, accuracy, and resource efficiency. This paper comprehensively examines YOLOv12's architectural innovations and their implications for real-time object detection. Following this introduction, we trace the evolutionary milestones of the YOLO family, setting the foundation for understanding how YOLOv12's core design elements area attention, R-ELAN, and 7$\times$7 separable convolutions collectively elevate the model's performance and expand its application horizon.


\section{Progression of YOLO Frameworks}

Table \ref{tab:yolo_versions} provides a comprehensive overview of the evolution of YOLO models, highlighting the key innovations and enhancements introduced with each iteration. These advancements have significantly strengthened object detection capabilities, improved computational efficiency, and expanded the versatility of the models to handle a wide array of computer vision tasks.

\begin{table}[ht]
\centering
\caption{YOLO: Evolution of Frameworks}
\label{tab:yolo_versions}
\begin{tabular}{|l|l|p{5cm}|p{5cm}|l|}
\hline
\textbf{Release} & \textbf{Year} & \textbf{Tasks} & \textbf{Contributions} & \textbf{Framework} \\ \hline
YOLO \cite{redmon2015you} & 2015 & Object Detection, Basic Classification & Single-stage object detector & Darknet \\
YOLOv2 \cite{redmon2016yolo9000betterfasterstronger} & 2016 & Object Detection, Enhanced Classification & Multi-scale training, dimension clustering & Darknet \\
YOLOv3 \cite{redmon2018yolov3} & 2018 & Object Detection, Multi-scale & SPP block, Darknet-53 backbone & Darknet \\
YOLOv4 \cite{bochkovskiy2020yolov4} & 2020 & Object Detection, Basic Tracking & Mish activation, CSPDarknet-53 backbone & Darknet \\
YOLOv5 \cite{yolov5_blog} & 2020 & Object Detection, Instance Segmentation & Anchor-free detection, SWISH activation, PANet & PyTorch \\
YOLOv6 \cite{li2022yolov6} & 2022 & Object Detection, Instance Segmentation & Self-attention, anchor-free OD & PyTorch \\
YOLOv7 \cite{wang2023yolov7} & 2022 & Object Detection, Tracking, Segmentation & Transformers, E-ELAN reparameterization & PyTorch \\
YOLOv8 \cite{Solawetz2023yolov8} & 2023 & Object Detection, Instance and Panoptic Segmentation & GANs, anchor-free detection & PyTorch \\
YOLOv9 \cite{wang2024yolov9} & 2024 & Object Detection, Instance Segmentation & PGI and GELAN & PyTorch \\
YOLOv10 \cite{wang2024yolov10} & 2024 & Object Detection & Consistent dual assignments for NMS-free training & PyTorch \\
YOLOv11 \cite{alif2024yolov11vehicledetectionadvancements} & 2024 & Object Detection, Instance Segmentation & Expanded capabilities, improved efficiency & PyTorch \\
\textbf{YOLOv12} \cite{tian2025yolov12attentioncentricrealtimeobject} 
& 2025 
& Object Detection, Instance Segmentation 
& Advanced attention-centric design (FlashAttention, R-ELAN), 7$\times$7 separable convolutions, improved computational efficiency 
& PyTorch \\ \hline
\end{tabular}
\end{table}

This progression demonstrates the steady evolution of real-time detection methodologies, with each version introducing novel techniques from the foundational single-stage detector of YOLO to increasingly sophisticated structures integrating self-attention and transformer-based components. YOLOv10 and YOLOv11 laid the groundwork for enhanced accuracy and efficiency in challenging scenarios, employing improved data augmentations and attention modules.

The most recent version, YOLOv12, builds on this legacy by introducing additional architectural refinements that further augment feature extraction and computational throughput. Specifically, YOLOv12 adopts an attention-centric design featuring FlashAttention, a novel Residual Efficient Layer Aggregation Network (R-ELAN), and 7$\times$7 separable convolutions, addressing the limitations of its predecessors. In the following sections, we will explore these advancements in detail, illustrating how YOLOv12 advances the state of the art in key computer vision tasks such as object detection and instance segmentation.

\section{YOLOv12: A Paradigm Shift in Real-Time Detection}

\begin{figure}[t]
    \centering
    \includegraphics[width=0.5\linewidth]{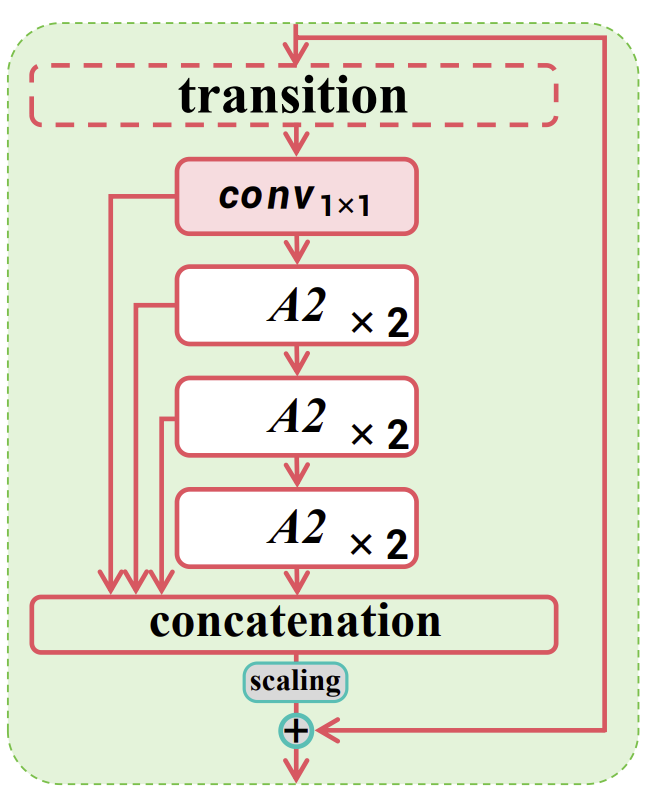}
    \caption{Residual Efficient Layer Aggregation Network as presented in paper \cite{tian2025yolov12attentioncentricrealtimeobject}}
    \label{fig:yolov12-rlean}
\end{figure}

YOLOv12 signifies a groundbreaking advancement in real-time object detection, representing a paradigm shift through the integration of attention-centric mechanisms, streamlined architectural designs, and optimised training pipelines. Building upon the robust foundations laid by its predecessors, YOLOv12 introduces a suite of enhancements aimed at maximising both accuracy and computational efficiency. At its core is a re-engineered feature extraction strategy that leverages the \emph{Residual Efficient Layer Aggregation Network} (R-ELAN), illustrated in Figure \ref{fig:yolov12-rlean}, \emph{FlashAttention}, and \emph{7$\times$7 separable convolutions} to deliver superior throughput and precision \cite{tian2025yolov12attentioncentricrealtimeobject}. By amalgamating these elements, YOLOv12 elevates performance in object detection and instance segmentation tasks, ensuring it can adeptly handle complex visual scenes with varying levels of detail and occlusion.

A hallmark of YOLOv12 lies in its adaptability to challenging detection scenarios. Through its refined \emph{area attention} module accelerated by FlashAttention, the model effectively isolates critical regions in cluttered or dynamic environments, enabling more accurate localisation of objects, including those that are small, partially obscured, or overlapping. Despite the inherent complexity of such tasks, YOLOv12 maintains the hallmark real-time processing speeds the series is celebrated for, making it especially suited for latency-sensitive applications like autonomous navigation and urban surveillance. In doing so, it sets new benchmarks for object detection performance and broadens the scope of what can be achieved in computer vision \cite{tian2025yolov12attentioncentricrealtimeobject}.

This latest iteration of YOLO underscores its versatility across diverse sectors. In the automotive domain, the enhanced detection capabilities of YOLOv12 can bolster advanced driver-assistance systems (ADAS) and self-driving vehicles, improving their ability to recognise critical road elements in real-time. In healthcare, YOLOv12's ability to handle small or complex objects may prove invaluable in analysing medical imaging data, while in agriculture, its streamlined processing pipeline allows on-the-fly monitoring of crops under constrained compute resources. By providing both efficiency and accuracy, YOLOv12 emerges as a pivotal tool for practitioners aiming to harness the full potential of real-time object detection technologies in practical, large-scale deployments.

\section{Architectural Blueprint of YOLOv12}

The success of the YOLO framework has historically rested on a unified architecture that performs bounding box regression and object classification, enabling fully differentiable, end-to-end training. YOLOv12 extends this core principle by integrating new architectural innovations tailored explicitly for higher accuracy, lower latency, and greater adaptability. As shown in Table \ref{tab:yolo12_structure}, the design of YOLOv12 can be divided into three main components: the \textbf{backbone}, which extracts and processes multi-scale features; the \textbf{neck}, which aggregates and refines those features; and the \textbf{head}, which generates the final predictions.

\begin{table}[ht]
\centering
\caption{Key Structural Components in YOLOv12}
\label{tab:yolo12_structure}
\begin{tabular}{|l|p{6cm}|p{6cm}|}
\hline
\textbf{Component} & \textbf{Functionality} & \textbf{Innovations in YOLOv12} \\ \hline
\textbf{Backbone} & Extracts features from input images at multiple scales using convolutional layers. & R-ELAN for deeper residual connectivity; 7$\times$7 separable convolutions to preserve spatial context with fewer parameters. \\ \hline
\textbf{Neck} & Aggregates features from different scales and transmits them to the head for predictions. & Area attention mechanisms powered by FlashAttention for efficient focus on critical regions. \\ \hline
\textbf{Head} & Generates final predictions, including bounding box coordinates and class labels. & Refined prediction pathways for accurate multi-scale detection; loss functions optimised for real-time performance. \\ \hline
\end{tabular}
\end{table}

\subsection{Backbone}
The backbone of YOLOv12 is crucial for converting raw image data into multi-scale feature maps, providing the foundational representations for subsequent detection tasks. Central to the backbone is the \emph{Residual Efficient Layer Aggregation Network} (R-ELAN), which fuses deeper convolutional layers with carefully placed residual connections. This design addresses gradient bottlenecks and enhances feature reuse, boosting the model's ability to capture intricate object details across various sizes and shapes.

\subsubsection{Advanced Convolutional Blocks}
Compared to earlier versions, YOLOv12 employs a new convolutional block class emphasising lightweight operations and higher parallelisation. These blocks utilise a series of smaller kernels, represented generically as:
\begin{equation}
F_{\text{out}} = \sum_{i=1}^{n} W_i * F_{\text{in}} + b_i,
\end{equation}
where $F_{\text{out}}$ is the output feature map, $W_i$ are the convolutional filters, $F_{\text{in}}$ is the input feature map, and $b_i$ is the bias term. By distributing the computation across multiple small convolutions instead of fewer large ones, YOLOv12 achieves faster processing without compromising feature extraction quality.

\subsection{Enhanced Backbone Architecture}
Beyond introducing advanced convolutional blocks, YOLOv12 leverages techniques like \emph{7$\times$7 separable convolutions} to reduce the computational burden. This approach effectively replaces conventional large-kernel operations or positional encodings, maintaining spatial awareness with fewer parameters. Additionally, multi-scale feature pyramids ensure that objects of varied sizes, including small or partially occluded ones, are represented distinctly within the network.

\subsection{Neck}
Functioning as a conduit between the backbone and head, the neck in YOLOv12 aggregates and refines multi-scale features. One of its key innovations is an \emph{area attention} mechanism accelerated by FlashAttention, which enhances the model's focus on critical regions in cluttered scenes. Mathematically, this can be interpreted as a segmented attention operation:
\begin{equation}
\text{Attention}(Q, K, V) = \text{softmax}\!\Big(\frac{QK^T}{\sqrt{d_k}}\Big)\,V,
\end{equation}
where \(Q, K, V\) are query, key, and value matrices, and \(d_k\) is the dimensionality of the key. By segmenting feature maps into areas and applying fast attention routines, YOLOv12 reduces memory transfers and computational overhead, enabling real-time inference even at higher input resolutions.

\subsection{Head}
The head of YOLOv12 transforms the refined feature maps from the neck into final predictions, generating bounding box coordinates and classification scores. Key improvements include streamlined multi-scale detection pathways, and specialised loss functions that better balance localisation and classification objectives. For example, a typical YOLO-style loss might be extended to incorporate new attention or confidence terms:
\begin{equation}
\mathcal{L} = \lambda_{\text{coord}} \sum (\hat{x} - x)^2 + (\hat{y} - y)^2 
\;+\; \lambda_{\text{obj}} \sum ( \hat{C} - C )^2 
\;+\; \ldots
\end{equation}
where \(\hat{x}, \hat{y}, \hat{C}\) denote predicted bounding box coordinates and confidence, respectively. Such refinements further enhance YOLOv12's performance in real-time applications.

YOLOv12 achieves a significant architectural evolution, blending innovative backbone elements, advanced attention mechanisms, and refined prediction modules. Together, these components set new standards for speed and accuracy in object detection while seamlessly extending to more specialised tasks such as instance segmentation.

\section{Core Computer Vision Tasks Facilitated by YOLOv12}

Designed to excel across a variety of computer vision challenges, YOLOv12 leverages its reimagined architecture and optimised algorithms to deliver robust performance in:
Designed to excel across a range of computer vision challenges, YOLOv12 leverages its reimagined architecture and optimised algorithms to deliver robust performance in:
\begin{enumerate}
    \item \textbf{Object Detection:} 
    Enhanced convolutional feature extraction and attention mechanisms enable precise localisation in real-time, ensuring high accuracy in applications such as autonomous vehicles and smart surveillance.

    \item \textbf{Instance Segmentation:} 
    By pairing its refined backbone with specialised segmentation heads, YOLOv12 partitions object at the pixel level, which is vital for domains like medical imaging and manufacturing defect detection.
\end{enumerate}

\begin{table}[t]
\centering
\caption{YOLOv12 Key Architectural Features}
\label{tab:yolo12_features}
\begin{tabular}{|p{3cm}|p{6cm}|p{6cm}|}
\hline
\textbf{Feature} & \textbf{Technical Details} & \textbf{Benefits} \\ \hline

\textbf{Enhanced Backbone Architecture} & 
- R-ELAN with deeper residual links. \newline
- 7$\times$7 separable convolutions for efficient spatial encoding. \newline
- Multi-scale feature pyramids.
&
- Improved capture of small or complex objects. \newline
- Faster, more accurate feature extraction. \newline
- Greater robustness under varied scene complexities. \\ \hline

\textbf{Advanced Attention Mechanisms} & 
- Area-based FlashAttention for reduced overhead. \newline
- Channel and spatial weighting for feature refinement. \newline
- Context-aware dynamic weighting.
&
- Strong focus on salient regions within images. \newline
- Enhanced detection in cluttered or dynamic environments. \newline
- Fewer false positives via targeted weighting. \\ \hline

\textbf{Optimized Neck Design} & 
- Enhanced feature aggregation strategies. \newline
- Depthwise separable layers to reduce compute cost. \newline
- Flexible up/down-sampling processes.
&
- Better multi-scale feature integration. \newline
- Lower computational overhead and faster inference. \newline
- Superior adaptability for different object sizes. \\ \hline

\textbf{Refined Head Modules} & 
- Larger receptive fields for improved contextual cues. \newline
- Non-linear activations (e.g., SiLU) are used to boost expressivity. \newline
- Fine-tuned bounding box regression.
&
- Higher precision in object localisation. \newline
- More robust object classification under varied conditions. \newline
- Smoother training convergence. \\ \hline

\textbf{Parameter Optimization} & 
- Lightweight convolutional blocks reduce trainable parameters. \newline
- Pruning and quantisation for edge deployment. \newline
- Streamlined architecture for memory efficiency.
&
- Smaller footprint for resource-limited devices. \newline
- Maintains competitive accuracy with fewer parameters. \newline
- Scalable performance across hardware. \\ \hline

\textbf{Enhanced Training Pipeline} & 
- Advanced data augmentations (Mosaic, MixUp). \newline
- Dynamic learning rate schedules and high-performing optimisers. \newline
- Transfer learning from large-scale datasets.
&
- Elevated model generalisation and robustness. \newline
- Faster convergence with diverse data distributions. \newline
- Consistent performance in real-world conditions. \\ \hline

\end{tabular}
\end{table}

As summarised in Table \ref{tab:yolo12_features}, each facet of YOLOv12's architecture and training pipeline is carefully tailored to deliver high-performing, efficient, and versatile solutions to modern computer vision challenges. By unifying profound architectural innovations with efficient attention mechanisms, YOLOv12 meets the high demands of real-time object detection. It expands its applicability to an ever-growing range of tasks and industries.

\section{Advancements and Key Features of YOLOv12}

YOLOv12 represents a significant leap forward in object detection, building upon the strong foundations laid by its predecessor, YOLOv11, introduced earlier in 2025. This latest iteration from Ultralytics leverages refined architectural designs, more sophisticated feature extraction techniques, and optimised training pipelines to maximise both speed and accuracy \cite{tian2025yolov12attentioncentricrealtimeobject}. Central to YOLOv12's improvements is its ability to detect subtle details in challenging scenarios thanks to advanced modules like the \emph{Residual Efficient Layer Aggregation Network} (R-ELAN), \emph{area attention} accelerated by FlashAttention, and \emph{7$\times$7 separable convolutions}. By integrating these innovations, YOLOv12 achieves a balanced synergy of rapid processing, high accuracy, and computational efficiency that positions it at the forefront of Ultralytics' model portfolio \cite{alif2024yolov11vehicledetectionadvancements}.

A key strength of YOLOv12 lies in its refined architecture, which targets a broader range of patterns and intricate elements within images. Compared to prior iterations, YOLOv12 introduces several notable enhancements:

\begin{enumerate}
    \item \textbf{Enhanced Precision with Optimized Complexity:} 
    The YOLOv12m variant achieves higher mean Average Precision (mAP) on the COCO dataset while using up to 25\% fewer parameters compared to YOLOv11m, highlighting the model's improved computational efficiency without sacrificing accuracy \cite{tian2025yolov12attentioncentricrealtimeobject}.

    \item \textbf{Expanded Versatility in CV Tasks:} 
    In addition to robust object detection, YOLOv12 supports instance segmentation via a re-engineered backbone (R-ELAN) and advanced neck (area attention), accommodating more complex pixel-level tasks for applications like medical imaging and industrial defect detection \cite{tian2025yolov12attentioncentricrealtimeobject}.

    \item \textbf{Optimized Speed and Performance:} 
    Through further refinement of convolutional blocks such as 7$\times$7 separable convolutions and a streamlined neck and head design, YOLOv12 achieves an optimal balance between latency and accuracy, making it well-suited for real-time scenarios \cite{tian2025yolov12attentioncentricrealtimeobject}.

    \item \textbf{Streamlined Parameter Count and Model Size:} 
    Reductions in parameter count lead to faster inference and lower memory consumption without significantly diminishing YOLOv12's detection quality. This efficiency is crucial for applications on resource-constrained hardware \cite{tian2025yolov12attentioncentricrealtimeobject}.

    \item \textbf{Advanced Feature Extraction Techniques:} 
    YOLOv12 integrates state-of-the-art improvements in the backbone (R-ELAN) and neck (area attention), enhancing feature extraction to address challenges such as small, partially occluded, or overlapping objects \cite{tian2025yolov12attentioncentricrealtimeobject}.

    \item \textbf{Contextual and Environmental Adaptability:} 
    By incorporating FlashAttention and dynamic multi-scale feature handling, YOLOv12 readily adapts to diverse deployment conditions from edge devices to large-scale cloud environments, ensuring robust performance under varied computational budgets \cite{tian2025yolov12attentioncentricrealtimeobject}.

    \item \textbf{Enhanced Training Methodologies:} 
    The training pipeline of YOLOv12 benefits from advanced data augmentation techniques (e.g., Mosaic, MixUp), dynamic learning rate schedules, and state-of-the-art optimisers. These refinements improve model generalisation and stability across heterogeneous datasets \cite{tian2025yolov12attentioncentricrealtimeobject}.
\end{enumerate}

\begin{figure}[t]
    \centering
    \includegraphics[width=1\linewidth]{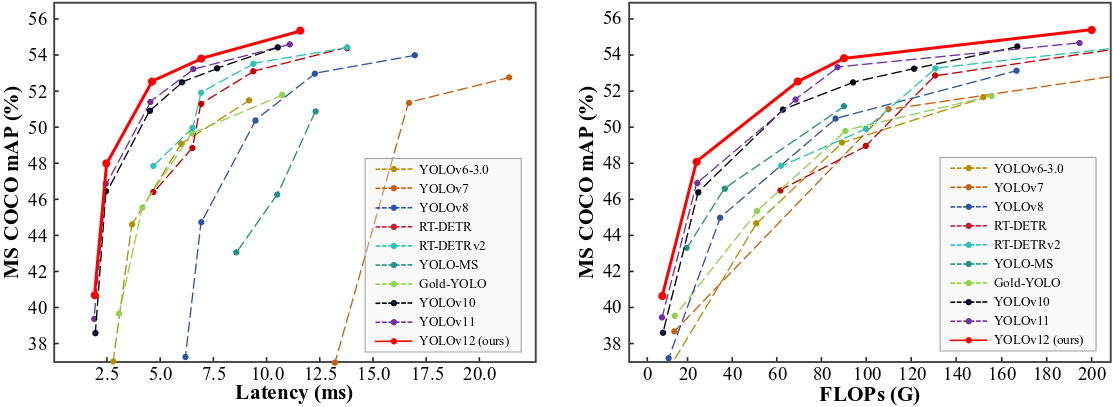}
    \caption{Benchmarking YOLOv12 Against Previous Versions \cite{tian2025yolov12attentioncentricrealtimeobject}}
    \label{fig:yolov12-vs-prev}
\end{figure}

In benchmark analyses, YOLOv12 consistently surpasses its predecessors, including YOLOv10 and YOLOv11, in both inference speed and accuracy. As depicted in Figure \ref{fig:yolov12-vs-prev}, YOLOv12 variants (\emph{12n}, \emph{12s}, \emph{12m}, and \emph{12x}) form a distinct performance frontier by achieving higher COCO mAP$^{50\text{-}95}$ at reduced latency points \cite{tian2025yolov12attentioncentricrealtimeobject}. Notably, \emph{YOLOv12x} reaches around 56\% mAP$^{50\text{-}95}$ at a mere 12ms inference time surpassing all prior YOLO versions. Smaller variants, such as \emph{YOLOv12m}, demonstrate exceptional efficiency by matching or exceeding larger models from previous generations while requiring significantly less processing time.

A significant breakthrough is observed in the low-latency regime (1–5ms), where \emph{YOLOv12s} maintains a high accuracy of approximately 49\% mAP$^{50\text{-}95}$. This performance tier was previously unattainable by models of similar speed, marking a milestone for real-time use cases that demand both speed and precision. Moreover, the scalability of YOLOv12 across its different variants indicates more efficient utilisation of additional computational resources compared to earlier YOLO generations.

As summarised in Table \ref{tab:yolo12_features}, each architectural feature in YOLOv12 has been carefully engineered to enhance speed, accuracy, and overall efficiency. By uniting advanced attention mechanisms with improved training methodologies, YOLOv12 delivers robust and versatile performance, solidifying its role as a preeminent choice for modern computer vision applications.

\section{Discussion}

YOLOv12 marks a substantial advancement in object detection technology, building on the strengths of YOLOv11 while incorporating novel architectural and algorithmic enhancements. Its core improvements revolve around increasing efficiency, broadening the scope of supported tasks, and maintaining real-time responsiveness, even under challenging conditions.

\begin{enumerate}
    \item \textbf{Scalability and Efficiency:} 
    YOLOv12 introduces multiple model variants (e.g., 12n, 12s, 12m, 12x) to accommodate diverse deployment settings. This tiered approach allows users to prioritise speed or accuracy depending on their application constraints. Smaller variants like 12n and 12s demonstrate significant gains in latency-sensitive tasks, making them excellent candidates for real-time embedded systems.

    \item \textbf{Architectural Innovations:} 
    With a redesigned backbone leveraging R-ELAN and incorporating 7$\times$7 separable convolutions, YOLOv12 significantly refines feature extraction and representation. These updates, combined with area-based attention accelerated by FlashAttention in the neck, result in faster processing without compromising accuracy. The enhanced backbone and neck architecture strengthen the model's ability to capture complex patterns, especially in cluttered scenes.

    \item \textbf{Instance Segmentation Capabilities:} 
    Beyond object detection, YOLOv12 readily adapts to instance segmentation by employing a shared backbone and specialised segmentation heads. This dual-task flexibility allows the model to tackle pixel-level object separation in domains such as medical imaging and manufacturing defect detection without incurring excessive computational overhead.

    \item \textbf{Attention-Centric Design:} 
    A notable leap from YOLOv11 is the integration of area attention mechanisms that harness FlashAttention to reduce memory overhead and improve focus on salient regions. This refined attention capability is critical for detecting smaller or partially occluded objects and contributes to YOLOv12's robust performance across various real-world scenarios.

    \item \textbf{Performance Highlights:} 
    Comparative benchmarks reveal consistent gains in mAP and inference speed across all YOLOv12 variants. In the low-latency regime, smaller models achieve accuracy levels previously out of reach for detectors operating at comparable speeds. Meanwhile, more significant variants maintain a high level of precision suitable for complex applications, illustrating the model's strong scalability.

    \item \textbf{Implications for Real-World Applications:} 
    The ability to perform robust detection at high frame rates broadens YOLOv12's applicability, benefiting use cases such as autonomous driving, where millisecond-level decisions can be critical, or real-time security systems needed to track fast-moving targets. Its reduced memory footprint and efficient processing make it amenable to deployment on edge devices, all while preserving high accuracy in challenging scenarios.

\end{enumerate}

In summary, YOLOv12 continues the YOLO tradition of driving real-time object detection forward through a carefully balanced mix of architectural refinements, attention enhancements, and parameter optimisations. The result is a flexible yet powerful model suite capable of handling diverse computer vision tasks under varying resource constraints. As industries and research domains increasingly focus on intelligent, time-critical applications, YOLOv12 stands poised to offer a practical, high-performance solution.

\section{Conclusion}

YOLOv12 marks a significant milestone in the evolution of real-time object detection, building on the established success of its predecessors while incorporating targeted architectural and algorithmic breakthroughs. By combining a more efficient backbone (R-ELAN), advanced attention modules powered by FlashAttention, and 7$\times$7 separable convolutions, YOLOv12 substantially enhances both speed and accuracy across a range of detection tasks. Moreover, its design readily adapts to instance segmentation, underscoring the model's versatility and potential for broader computer vision applications.

Empirical results demonstrate that YOLOv12 consistently achieves higher mAP and faster inference speeds than earlier YOLO variants, making it an appealing choice for time-sensitive applications such as autonomous driving, security surveillance, and real-time analytics. The adaptability and scalability of YOLOv12 allow for deployment on resource-constrained edge devices and high-performance GPU clusters, underscoring its versatility across diverse operational environments. Additionally, as shown by various lightweight CNN and attention-based methods \cite{boltAttention, boltVision, palletRacking, retinopathyExudate, helmetSafety, banglaResnet, banglaCNN}, the future of deep learning lies in balancing efficiency with robust performance, an ethos that YOLOv12 reinforces at scale.

Overall, YOLOv12 advances the frontier of real-time object detection by striking an optimal balance between computational efficiency and state-of-the-art performance. Its novel architectural refinements and training optimisations position it as a robust solution for modern computer vision challenges, paving the way for further innovation in research and industrial applications.

\vspace{6pt} 


\bibliographystyle{unsrt}  
\bibliography{ref}  

\end{document}